\documentclass[letterpaper]{article}

\usepackage{natbib,alifeconf}  
\usepackage{amsfonts}
\usepackage{amsmath}
\usepackage{wrapfig}
\usepackage{bbold}
\usepackage{hyperref}
\usepackage{subcaption}
\usepackage{stackengine}
\setstackgap{L}{\normalbaselineskip}
\strutlongstacks{T}

\title{Flow-Lenia: Towards open-ended evolution in cellular automata through mass conservation and parameter localization}
\author{Erwan Plantec$^1$, Gautier Hamon$^1$, Mayalen Etcheverry$^{1, 2}$, Pierre-Yves Oudeyer$^1$,\\ {\Large Clément Moulin-Frier$^1$ and Bert Wang-Chak Chan$^{1, 3}$} \\
\mbox{}\\
$^1$FLOWERS Team, Inria, Bordeaux, France \\
$^2$Poietis, Pessac, France \\
$^3$Brain Team, Google research, Tokyo, Japan\\
eplantec@gmail.com} 

\begin{document}

\maketitle


\begin{abstract}
  \newcommand{\website}{\url{https://sites.google.com/view/flowlenia/videos}}
\newcommand{\notebook}{\url{https://tinyurl.com/mr2ncy3h}}

The design of complex self-organising systems producing life-like phenomena, such as the open-ended evolution of virtual creatures, is one of the main goals of artificial life. Lenia, a family of cellular automata (CA) generalizing Conway's Game of Life to continuous space, time and states, has attracted a lot of attention because of the wide diversity of self-organizing patterns it can generate. Among those, some spatially localized patterns (SLPs) resemble life-like artificial creatures and display complex behaviors. However, those creatures are found in only a small subspace of the Lenia parameter space and are not trivial to discover, necessitating advanced search algorithms. Furthermore, each of these creatures exist only in worlds governed by specific update rules and thus cannot interact in the same one. This paper proposes as mass-conservative extension of Lenia, called Flow Lenia, that solve both of these issues. We present experiments demonstrating its effectiveness in generating SLPs with complex behaviors and show that the update rule parameters can be optimized to generate SLPs showing behaviors of interest. Finally, we show that Flow Lenia enables the integration of the parameters of the CA update rules within the CA dynamics, making them dynamic and localized, allowing for multi-species simulations, with locally coherent update rules that define properties of the emerging creatures, and that can be mixed with neighbouring rules. We argue that this paves the way for the intrinsic evolution of self-organized artificial life forms within continuous CAs. A notebook with Flow Lenia implementation and demo are available at \notebook.
\end{abstract}

\newcommand{\R}{\mathbb{R}}
\newcommand{\N}{\mathbb{N}}
\newcommand{\supp}{\mathcal{L}}
\newcommand{\website}{\url{https://sites.google.com/view/flowlenia/videos}}
\newcommand{\notebook}{\url{https://colab.research.google.com/drive/1l-Og8xRlc5ew0489swuud0Me7Sc5bCss?usp=sharing}}

\section{Introduction}

Complex self-organizing systems have been central to Artificial Life (ALife) towards the search of emergent life-reminiscent phenomenons. Among those systems are continuous cellular automata (CAs) like SmoothLife \citep{rafler2011} or Lenia \citep{chan2019}. Lenia is a family of CAs generalizing Conway's Game of Life (GoL) to continuous space, time and states \citep{chan2019,chan2020}. Each instance of Lenia is given by a specific configuration of parameters defining its update rule and an initial configuration which map to a specific pattern and behavior in a deterministic way. Lenia is a generalization of GoL as the latter corresponds to a specific parameter set in Lenia. 

\begin{figure}[ht!]
\begin{tabular}{ccc}
     \subfloat[]{\includegraphics[width=.27\columnwidth]{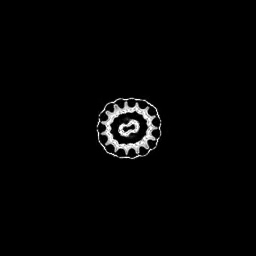}}&  
     \subfloat[]{\includegraphics[width=.27\columnwidth]{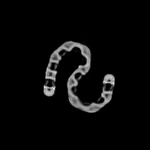}}&
     \subfloat[]{\includegraphics[width=.27\columnwidth]{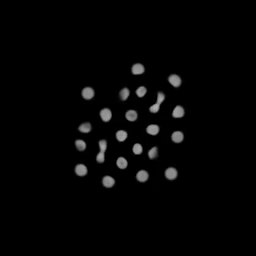}}\\
     \subfloat[]{\includegraphics[width=.27\columnwidth]{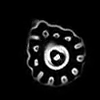}}&  
     \subfloat[]{\includegraphics[width=.27\columnwidth]{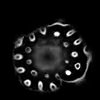}}&
     \subfloat[]{\includegraphics[width=.27\columnwidth]{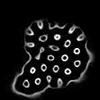}}\\
     \subfloat[]{\includegraphics[width=.27\columnwidth]{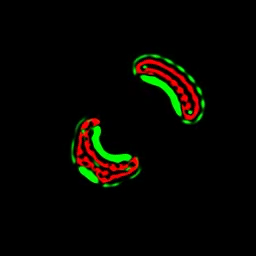}}&  
     \subfloat[]{\includegraphics[width=.27\columnwidth]{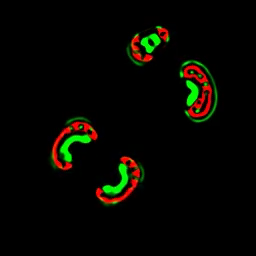}}&
     \subfloat[]{\includegraphics[width=.27\columnwidth]{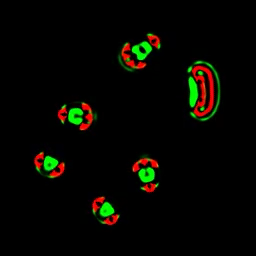}}\\
     \multicolumn{3}{c}{\subfloat[]{\includegraphics[width=.9\columnwidth]{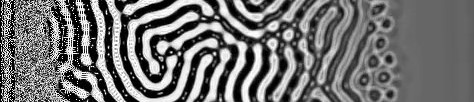}}}
\end{tabular} 
\caption{Flow Lenia creatures. (a-c) Samples of creatures found through random search in Flow Lenia parameter space. (d-f) and (g-i) Timelapses of patterns found through random search. Colors in (g-i) code for different channels. (j) Effect of changing temperature in Flow Lenia, temperature is linearly increasing from left to right. Videos are available at \website.}
\label{fig:random-search}
\end{figure}

Previous studies with Lenia have shown the emergence of autopoïetic (i.e self-produced) spatially localized patterns (SLPs), also called ``creatures'', often resembling microscopic life-forms and displaying various behaviors like motility or self-replication. Such observations have made Lenia a particularly interesting system for studying the emergence of life-like phenomenon and even sensori-motor capabilities \citep{hamon:hal-03519319}. Importantly, these artificial creatures bear a deep similarity to their biological counterparts as enacted agents endowed with constitutive autonomy (i.e creatures are self-producing) as described by enactive theories of cognition (see \citet{froese2009} for a more complete discussion on enactive artificial intelligence). \\
However, those patterns are quite difficult to find, necessitating complex fine-tuned search algorithms in order to find update rule parameters allowing the emergence of interesting patterns (e.g spatially localized patterns). For instance, complex patterns have been found in \citep{hamon:hal-03519319} using intrinsically motivated goal exploration processes and gradient descent. \\
An other important challenge in ALife and artificial intelligence is about the design of systems displaying open-ended intrinsic evolution (i.e unbounded growth of complexity through intrinsic evolutionary processes) \citep{stanley2019}. Such a process is called \emph{intrinsic} since no final objective (i.e fixed fitness function) is set by the experimenter, the fitness landscape is intrinsic to the system and depends only on its current state, as in natural evolution where there is no final goal \citep{lehman2011a}. Systems like EvoLoop \citep{sayama1999} or an extension of it, SexyLoop \citep{oros2007}, allow for evolutionary processes to occur in CAs. However, such systems rely on hand-defined rules and particular structures ultimately limiting the kind of patterns that can emerge in the system. On the other hand, even though Lenia creatures display greater diversity, different creatures (governed by different update rules) cannot exist in the same world (i.e the same simulation) and thus cannot interact. Obtaining such an evolutionary process in a CA could be achieved by embedding information in the system locally modifying the update rule and so the properties of emerging creatures, like a genome, enabling multi-species simulations. Such simulations might set the stage for evolution to occur in populations of patterns each with their own update rule. However, achieving it in CAs like Lenia is still an open-problem. \\
We believe that adding mass conservation is a key ingredient to address the aforementioned challenges. Such a constraint could (i) constrain emerging creatures to spatially localized ones, (ii) allow for the design of multi-species simulations and (iii) provide an important evolutionary pressure \citep{hickinbotham2015}. We propose in this work a mass-conservative extension to Lenia called \emph{Flow Lenia} and demonstrate that such conservation laws effectively facilitate the search fort artificial creatures by constraining (almost all) emerging patterns to spatially localized ones. We also show that the update rule parameters can easily be optimized using vanilla evolutionary strategies \citep{salimans2017} with respect to some fitness functions to obtain patterns with specific properties such as directed motion or angular motion. Importantly, we show that the Flow Lenia formulation enables the integration of the parameters of the CA update rules within the CA dynamics, making them dynamic and localized, allowing for multi-species simulations, with locally coherent update rules that define properties of the emerging creatures. Finally, we propose a resource collection mechanism which could act as a selective pressure in such multi-species simulations. 

\section{Model}

Let $\supp$ be the support of a CA, which is the two-dimensional grid $\mathbb{Z}^2$ in the rest of this work. Let $A^t: \supp \to S^C$ be CA's activations at time $t$, with $A^t_i(x)$ the activation in location $x \in \supp$, channel $i$ and time $t$. $C$ is the number of channels of the system ans $S$ is the state space (the set of states a cell can take in each channel).

\begin{figure*}[ht!]
    \centering
    \includegraphics[width=.9\textwidth]{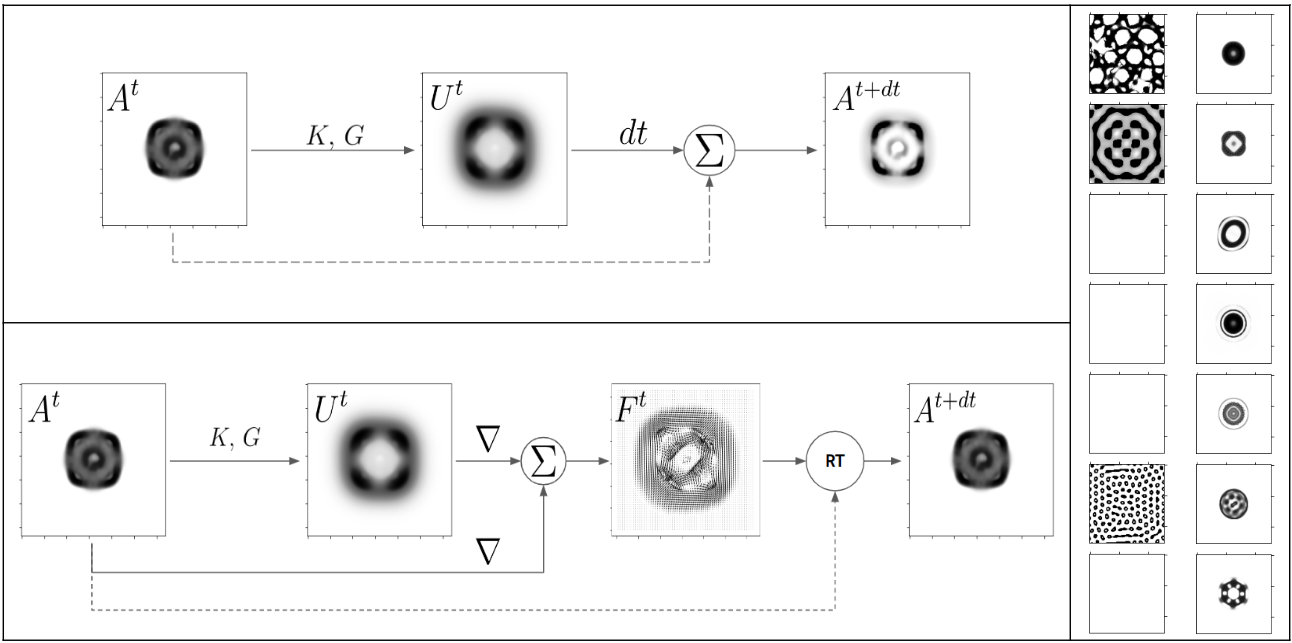}
    \caption{Top left : Lenia update rule. The growth $U^t$ is computed with kernels $K$ and growth functions $G$ (equation \ref{PotEq}) defined by a specific parameter configuration sampled in Lenia's parameter space. A small portion of the growth is then added to activations $A^t$ to give the next state $A^{t+dt}$ (equation \ref{equ:lenia_update}). Bottom left : Flow Lenia update rule. Affinity map $U^t$ is computed as in Lenia. The flow $F^t$ is given by combining the affinity map and concentration (i.e activations) gradients (equation \ref{FlowEquation}). Finally, the next state is obtained by ``moving'' matter in the CA space according to the flow $F^t$ using reintegration tracking (RT) (equations \ref{FlowApplyEq} and \ref{MatterFlowEq}). Right : Patterns generated from randomly sampled parameters sets (no cherry picking) in Lenia (left) and Flow Lenia (right). Parameters used for both systems are the same (i.e each row corresponds to one parameters set).}
\label{fig:lenia-flowlenia}
\end{figure*}

\subsection{Lenia}

The state space $S$ in Lenia is the unit range $[0, 1]$. An instance of Lenia is defined by a tuple $<K, G, A^0>$ where $K$ is a set of convolution kernels where $K_i: \mathcal{L} \to [0, 1]$ satisfies $\int_{\mathcal{L}}K_i = 1$ and $G$ is a set of growth functions with $G_i: [0,1] \to [-1,1]$. Each pair $(K_i, G_i)$ is associated to a source channel $c^i_0$ it senses and a target channel $c^i_1$ it updates. Connectivity can be represented through a square adjacency matrix $M_{C, C} = \left[ \begin{array}{ccc} m_{11} & \cdots & m_{1C} \\ \vdots & \ddots & \vdots \\ m_{C1} & \cdots & m_{CC} \end{array}\right]$ where $m_{ij} \in \N$ is the number of kernels sensing channel $i$ and updating channel $j$. $A^0$ is the initial state of the system.
As in \citet{hamon:hal-03519319}, kernels are radially symmetrical and defined as a sum of concentric Gaussian bumps :
\begin{equation} \label{equ:kernel}
        K_i(x) = \sum_{j=1}^{k} b_{i, j} \, exp\left(- \frac{(\frac{x}{r_iR} - a_{i, j})^2}{2 w_{i, j}^2}\right) 
\end{equation}
Where $a_i$, $b_i$, $w_i$ and $r_i$ are parameters defining kernel $i$. $k$ is a parameter defining the number of rings per kernel (set to 3 here) and $R$ is a parameter common to all kernels defining the maximum neighborhood radius. Each kernel is then defined by $3\times k + 1$ parameters.
Growth functions are defined as Gaussian function scaled in the range $[-1, 1]$:
\begin{equation} \label{equ:growth_function}
        G_i(x) = 2 \; exp\left(- \frac{(\mu_i - x)^2}{2\sigma_i^2}\right) - 1
\end{equation}
Where $\mu_i$ and $\sigma_i$ are parameters of growth function $i$ so each growth function is defined by $2$ parameters.
A step in Lenia is defined by the following steps (see figure \ref{fig:lenia-flowlenia} (top)) :
\begin{enumerate}
    \item Compute the growth at time $t$ given the actual state $A^t$ :
        \begin{equation} \label{PotEq}
            U^t_j = \sum_{i} h_i \cdot G_i(K_i \ast A_{c^i_0}^t) \mathbb{1}_{[c^i_1 = j]}
        \end{equation}
        Where $h \in \mathbb{R}^{|K|}$ is a vector weighting the impact of each pair $(K_i, G_i)$ on the growth.
    \item Add a small portion of the growth $U^t$ to the actual state $A^t$ to get the state at the next time step and clip results back to the unit range :
        \begin{equation} \label{equ:lenia_update}
            A^{t+dt}_i = [A^t_i + dt \, U^t_i]_0^1
        \end{equation}
\end{enumerate}

\subsection{Flow Lenia}

Flow Lenia is a mass-conservative extension to Lenia. By mass-conservative, we mean that the sum of activations across all cells and for each channel is constant over time :
\begin{equation*}
    \sum_{x\in \mathcal{L}} A^t_c = \sum_{x\in \mathcal{L}} A^{t+dt}_c \;, \forall t, \forall c \in \{1, ..., C\}
\end{equation*}

\begin{figure}
\centering
\includegraphics[scale=0.28]{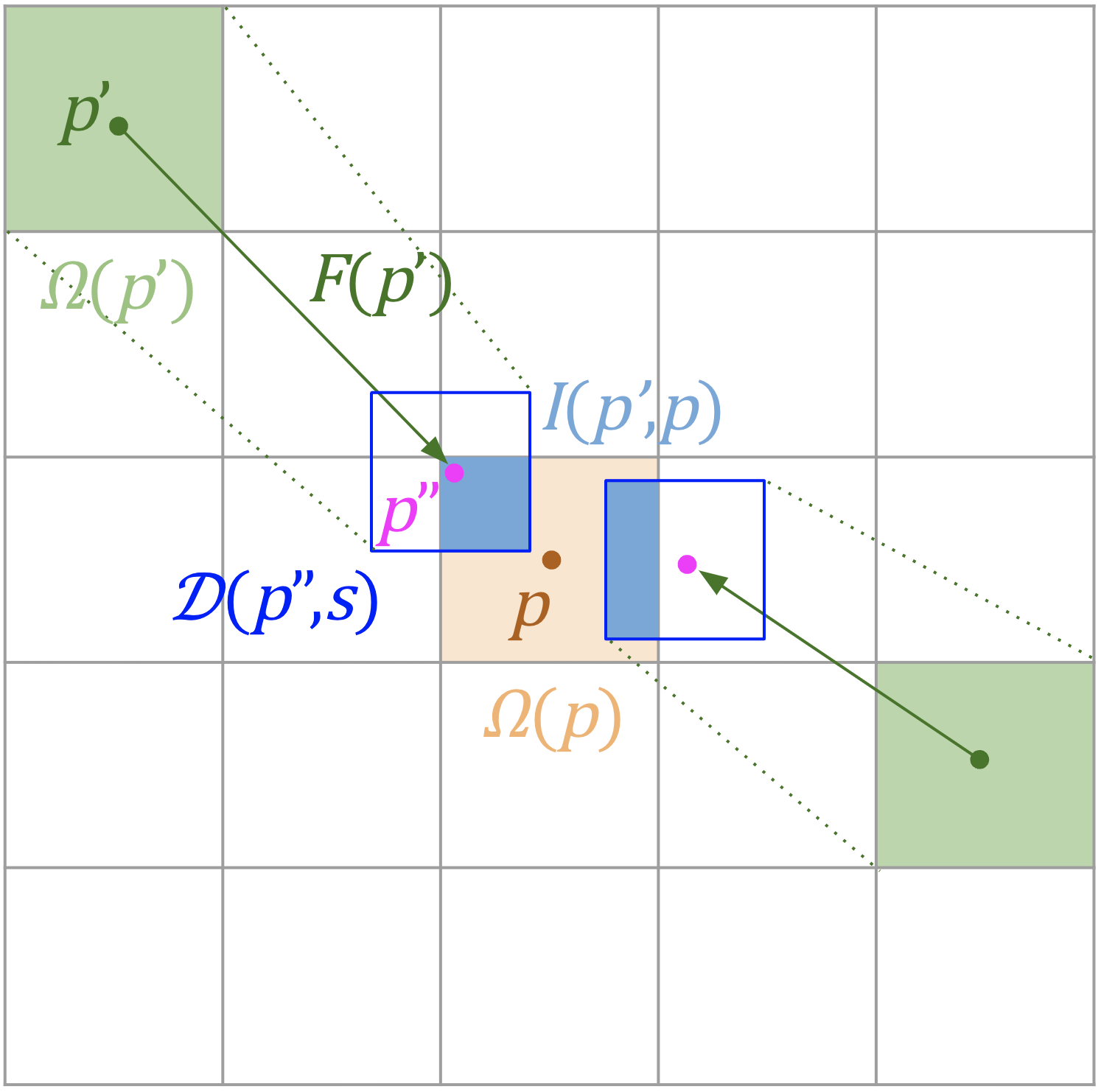}
\caption{Calculation of incoming matter to cell $p \in \mathcal{L}$ through reintegration tracking \citep{moroz2020}. Mass contained in cell at location $p' \in \mathcal{L}$ is moved to a square distribution $\mathcal{D}$ centered on $p'' = p' + dt \cdot F^t(p')$. The proportion of mass from $p'$ arriving in $p$ is then given by the integral of $\mathcal{D}$ on the cell domain of $p$, $\Omega(p)$, denoted as $I(p', p)$.}
\label{fig:fig-flow}
\end{figure}

We propose for this system to interpret activations as concentrations of ``matter" in all cells and to refer to the term $U^t$, previously called the growth in Lenia, as an affinity map. The idea is that the matter will move towards higher affinity regions by following the local gradient of the affinity map $U$, $\nabla U : \supp \to \R^2$. To do so, we define a flow $F : \mathcal{L} \to (\mathbb{R}^2)^C$, which can be interpreted as the instantaneous speed of matter, as:
\begin{equation} \label{FlowEquation}
    \begin{cases}
        F^t_i = (1 - \alpha^t) \nabla U^t_i - \alpha^t \nabla A^t_{\Sigma} \\
        \alpha^t(p) = [({A^t_{\Sigma}(p)} / {\theta_A})^n]_0^1
    \end{cases}
\end{equation}
With $A^t_{\Sigma}(p) = \sum_{i=1}^C A^t_i(p)$ the total mass in each location $p$. Here $\nabla U^t_i$ is the affinity gradient for channel $i$. The negative concentration gradient $- \nabla A^t_{\Sigma}$ is a diffusion term to avoid concentrating all the matter in very small regions akin to the clipping in Lenia which upper bounds concentrations. In practice, gradients are estimated through Sobel filtering. Map $\alpha : \mathcal{L} \to [0, 1]$ is used to weight the importance of each term such that $-\nabla A^t_{\Sigma}$ dominates when the total mass at a given location is close to a critical mass $\theta_A \in \mathbb{R}_{>0}$. Intuitively, the result is that matter is mainly driven by concentration gradients in high concentrations regions and is more free to move along the affinity gradient in less concentrated areas. We typically use $n>1$ such that the affinity gradient dominates on a larger range of masses. \\
Then, we can move matter in space according to flow $F$ giving us the state at the next time step. To do so we use the reintegration tracking method proposed in \citep{moroz2020}. Reintegration tracking is a semi-Lagrangian grid based algorithm thought as a reformulation of particle tracking in screen space (i.e grid space) aimed at not losing information (i.e particles) which happens when two particles end up in the same cell. The basic principle is to work with distribution of particles (i.e infinite number of particles) and conserve the total mass by adding up masses going on a same cell. Overall, reintegration tracking can be seen as a grid-based approximation to particle systems with infinite number of particles having the property to conserve total mass. Thus, Flow lenia can be seen as a new kind of model at the frontier between continuous CAs and particle systems. A particle based model directly inspired by the Flow Lenia formulation has been recently proposed in \citet{mordvintsev2022}. Figure \ref{fig:fig-flow} illustrates how reintegration tracking is used in our case. The resulting update rule is the following :
\begin{equation} \label{FlowApplyEq}
    A^{t+dt}_i(p) = \sum_{p' \in \mathcal{L}} A^{t}_i(p') I_i(p', p) 
\end{equation}
Where $I_i(p', p)$ is the proportion of incoming matter in channel $i$ going from cell $p' \in \mathcal{L}$ to cell $p \in \mathcal{L}$:
\begin{equation} \label{MatterFlowEq}
    I_i(p', p) = \int_{\Omega(p)} \mathcal{D}(p_i'', s) 
\end{equation}
With $p_i'' = p' + dt \cdot F^t_i(p')$ the target location of the flow from $p'$. $\Omega(p)$ is the domain of cell at location $p$, which is a square of side $1$. $\mathcal{D}(m, s)$ is a distribution defined on $\mathcal{L}$ with mean $m$ and variance $s$ satisfying $\int_{\mathcal{L}} \mathcal{D}(m, s) = 1$, which is in practice a uniform square distribution with side length $2s$ centered at $m$. This distribution emulates a flow of particles from source area $\Omega(p')$ to target area $\mathcal{D}(p'', s)$, where the distribution $\mathcal{D}$ emulates Brownian motion at the low level. $s$ is an hyperparameter of the system which can be seen as form of temperature. The reintegration tracking method is depicted in Fig. \ref{fig:fig-flow}. Since the distribution $\mathcal{D}$ integrates to 1, it is clear that a cell cannot send out more mass than it contains nor less and so the system conserves its total mass. Mass conservation also implies that cells' states are no longer bound to the unit range but can be any positive real valued number ($S \equiv \mathbb{R}_{\geq 0}^C$). This model has been implemented in JAX \citep{jax2018github} allowing fast simulation on GPU ($255 \mu s \pm 3.11 \mu s$ per step on Tesla T4 GPU with 1 channel, 10 kernels and $128 \times 128$ world size).

\section{Results}

\begin{table}[ht]
    \centering
    \begin{tabular}{|c c c | c c c|}
        \hline
        \multicolumn{3}{|c|}{\textbf{Neighborhood}} & \multicolumn{3}{|c|}{\textbf{Growth functions}}\\
        \hline
        $R$ &  $\in [2, 25]$ & & $\mu$ &   $\in [0.05, 0.5]$ & *\\
        $r$ &  $\in [0.2, 1]$ & * & $\sigma$ &  $\in [0.001, 0.2]$ & * \\
        \hline 
        \multicolumn{3}{|c|}{\textbf{Kernels }} & \multicolumn{3}{|c|}{\textbf{Flow }}\\
        \hline
        $h$ &  $\in [0, 1]$ & * & $s$ &  $0.65$ & \\
        $a$ &  $\in [0, 1]^3$ & * & $n$ &  $2$ & \\
        $b$ &  $ \in [0, 1]^3$ & * & $dt$ &  $0.2$ & \\
        $w$ &  $\in [0.01, 0.5]^3$ & * & & & \\
        \hline 
         
    \end{tabular}
    \caption{Flow Lenia explored parameter space. Parameters marked with a * must be sampled for each kernel-growth function pair.}
    \label{tab:flow_params}
\end{table}

\subsection{Random search}

By performing random and manual search of the Flow Lenia parameter and hyperparameter space described in table \ref{tab:flow_params} we have been able to discover SLPs with already interesting and complex behaviors. Such patterns are shown in figure \ref{fig:random-search} and described hereafter. Initial patterns $A^0$ are set with a $40\times 40$ patch with matter drawn from uniform distribution in the center of the grid and no matter everywhere else.\\
Most of the patterns generated in Flow Lenia are SLPs (see figure \ref{fig:lenia-flowlenia}) with rare exceptions found by manually setting parameters to specific configurations leading to scattered matter. Using multiple kernels led to the emergence of SLPs with more complex shapes and behaviors. While part of emerging patterns tend to be static ones, dynamic patterns are quite common in Flow Lenia. For instance gyrating SLPs (a) or snake like patterns (b) with complex motion emerging from attraction/repulsion dynamics can be frequently observed. Dividing and merging dots (c) resembling reaction-diffusion patterns are also a common pattern. Timelapse (d-f) shows a creature with complex and unpredictable dynamics emerging from the interactions of its membrane, multiple organoids-like structures and a central nuclei ultimately leading to a phase transition happening in (e). Timelapse (g-e) shows a 2-channels creature displaying complex division patterns and interesting modular creatures whose characteristics change depending on their total mass while being of the same ``kind" (i) (see 5 creatures on the leftmost part of (i)). Note that multi-channel creatures often show more complex dynamics and patterns with very modular shapes where each channel seems to occupy a different role. (j) shows the effect of changing the size of the reintegration tracking distribution $s$ (see equation \ref{MatterFlowEq} and figure \ref{fig:fig-flow}), parameter we call temperature. Here temperature is linearly increasing from left to right showing very different phases of the systems. More interestingly, patterns at the frontier between the Turing-like phase (center) and the equilibrium phase (right) are much more dynamic and display unpredictable dynamics suggesting a critical regime.

\subsection{Optimizing Flow Lenia creatures}

\begin{figure}[ht]
\centering
\includegraphics[width=\columnwidth]{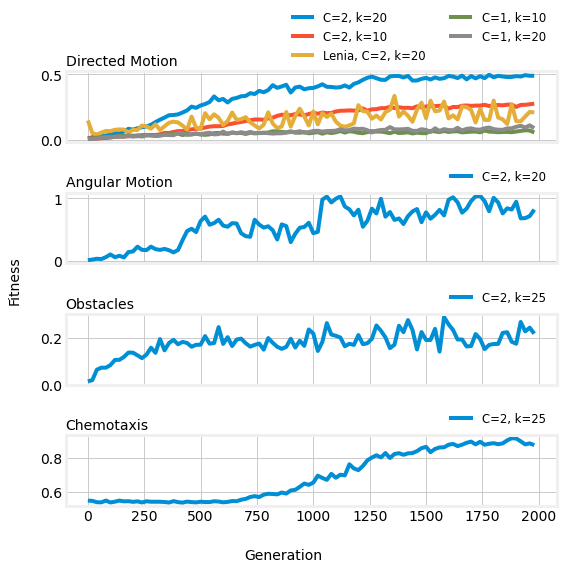}
\caption{Results of evolutionary optimization. C is the number of channels of the system and $k$ is the number of kernels and growth functions. When performing the exact same optimization for directed motion in the original Lenia system (yellow curve), not only optimization is unstable but it only discovers exploding patterns.}
\label{fig:evo_results}
\end{figure}

\begin{figure}[t]
\centering
\begin{tabular}{cc}
     \subfloat[]{\includegraphics[width=.4\columnwidth]{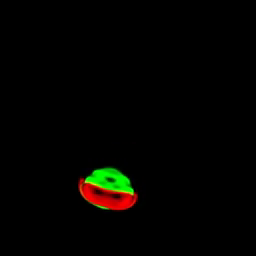}} &
     \subfloat[]{\includegraphics[width=.4\columnwidth]{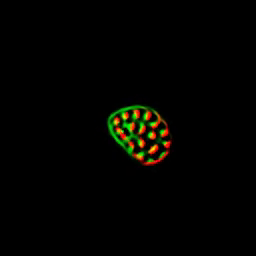}} \\
     \subfloat[]{\includegraphics[width=.4\columnwidth]{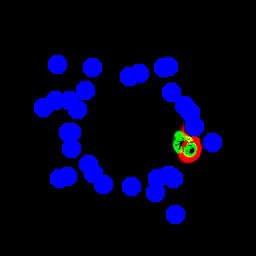}} &
     \subfloat[]{\includegraphics[width=.4\columnwidth]{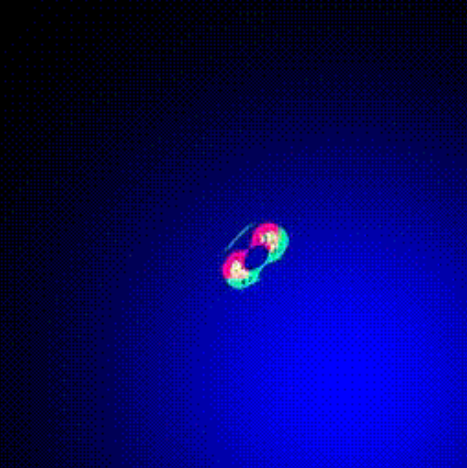}} \\
\end{tabular}
\caption{Creatures found through optimization. (a) Directed motion with 2 channels and 20 kernels. (b) Angular motion with 2 channels and 20 kernels. (c) Motion through obstacles with 2 channels and 25 kernels. (d) Chemotaxis with 2 channels and 25 kernels. Videos ara availabe at \website.}
\label{fig:evo_creas}
\end{figure}

In this section, we show that Flow Lenia update rule parameters can also be easily optimized so to generate patterns with specific behaviors. This is a difficult task in Lenia as it would require constantly monitoring the existential status and the spatially-localizedness of evolved creatures. Thus training creatures in Lenia requires to define characterizations of creatures accounting for such properties which is a far from trivial problem. Moreover, even if one can come up with proxies to find spatially localized patterns, the optimization process remains difficult necessitating advanced optimization methods like curriculum learning \citep{hamon:hal-03519319}. In Flow Lenia, the spatial localization constraint is intrinsic to the system thus removing the necessity to account for it when searching for creatures. Using evolutionary strategies \citep{salimans2017} to optimize the update rule parameters and the initial configuration ($A^0$) with respect to user-defined fitness functions, we have been able to successfully find good solutions for 4 different tasks : directed motion, angular motion, navigation through obstacles and chemotaxis.\\ 
We used evosax \citep{lange2022} implementation of the OpenES strategy with population size of $16$ and adam optimizer \citep{kingma2017} with $0.01$ as learning rate. We optimized the Flow Lenia update rule with different number of kernels and either 1 or 2 channels. For comparison, we also trained original Lenia on the directed motion task following the same optimization procedure. The initial pattern is composed, as in random search, of a square patch with non-zero activations placed at the center of the world and zeros everywhere else. Results are shown in figure \ref{fig:evo_results}. 

\subsubsection{Directed motion}
In order to train creatures displaying directed motion, i.e straight line motion, we used the distance travelled by the creature as the fitness function. The distance is calculated by computing the center of mass of the pattern at step 0 and final step 400. Formally, the fitness function is defined as :

\begin{equation*}
    f(\theta) = dist(\phi(A^0), \phi(A^{400}))
\end{equation*}

Where $A \equiv \{A^0, ..., A^{T}\}$ is the pattern obtained by making a rollout with parameters $\theta$ for $T$ timesteps (here 500). $\phi(A^t) \in [-0.5, 0.5]^2$ is the center of mass of state $A^t$ and $dist$ is the euclidean distance function.
We optimized the system with either 1 or 2 channels and 10 or 20 kernels. We used $M = \left[\begin{array}{cc}5 & 5 \\ 5 & 5\end{array}\right]$ as the adjacency matrix with 2 channels and 20 kernels and $M = \left[\begin{array}{cc}3 & 2 \\ 2 & 3\end{array}\right]$ with 10 kernels.\\
Results (see fig \ref{fig:evo_results}) show that good solutions can be found in the 2 channels condition but not in the single channel case. However, when running the algorithm for longer (e.g 5000 generations), we have been able to found single channel creatures with similar fitness than their 2 channels counterpart. Increasing the number of kernels led to faster discovery of good solutions. The best performing creature is shown in figure \ref{fig:evo_creas}(a). This creature moves because of attraction/repulsion dynamics between the 2 channels which might explain why directed motion is much easier to attain with multi-channels creatures. On the other hand, the optimization of the original Lenia model is much less stable and discovered patterns are less successful than their mass-conservative counterparts. Moreover, every Lenia optimized patterns are exploding ones.
\subsubsection{Angular motion}
In this task, we want emerging creatures to display more complex forms of motion. More precisely, we want creatures to be able to move and make turns. As with directed motion, we use the center of mass of the creature through time to compute its trajectory. The fitness function is the following :

\begin{align*}
    f(\theta) & = dist(\phi(A^0), \phi(A^{200})) \\
    & + dist(\phi(A^{200}), \phi(A^{400})) \\ 
    & + \angle [\phi(A^{200}) - \phi(A^{0})] \, [\phi(A^{400}) - \phi(A^{200})]
\end{align*}

Where $\angle a b$ is the angle between vectors $a$ and $b$. The first two terms are the distance travelled from step 0 to step 200 and from step 200 to step 400. The last term is the angle between these two trajectories which is maximal when they are opposite. In order to avoid large angles to come from very small movements, the angle is set to 0 when distance traveled either before or after step 200 is below a given threshold. The optimal behavior for this fitness function is then to move fast in one direction, make a 180° turn, and then move fast in the opposite direction. We used 2 channels, 20 kernels and the same connectivity matrix as for directed motion.\\ 
Result are shown in figure \ref{fig:evo_results}. The best performing creature, shown in figure \ref{fig:evo_creas} (b), displays very complex internal dynamics leading it to periodically make 180° turns while moving in straight line the rest of the time. These dynamics seem to be generated by attraction repulsion dynamics like the ones observed in directed motion but here in a more intricate morphology.

\subsubsection{Navigation through obstacles}

In this task, we want to see if creatures can navigate through obstacles as done in \citet{hamon:hal-03519319}. To do so, we added walls which are implemented by adding a strong flow going from the center of walls outwards, thus strongly repelling the creature and acting as a solid obstacle. At each evaluation of the optimization process, we randomly sample points on a circle surrounding the creatures' initial positions to be walls positions thus making a ``forest" of walls around the creature. We then optimize the creature with the same fitness function as in the \emph{directed motion} task so creatures have to go as far as possible and so through the forest. We made the experiment with 2 channels creatures, walls are defined in a separate third channel. We used 25 kernels and $M = \left[\begin{array}{ccc}5 & 5 & 0 \\ 5 & 5 & 0 \\ 5 & 0 & 0\end{array}\right]$ as the connectivity matrix so creatures are able to sense the walls channel (3rd channel).\\ 
We have been able to successfully train creatures able to move and stay robust when making contact with walls such as the one shown in figure \ref{fig:evo_creas} (c) which is able to resist deformation and find a way out the ``forest". In comparison, solving a similar task in Lenia required complex optimization methods based on curriculum learning, diversity search and gradient descent over a differentiable CA \citep{hamon:hal-03519319}. However, such a comparison is difficult because Flow Lenia creatures are inherently more robust due to conservation of mass, whereas Lenia creatures can disappear because of perturbations.

\subsubsection{Chemotaxis}

Another important feature of natural life-forms is the ability to sense their environment in order to find food or avoid dangers through chemotaxis. In this task, we want creatures to be able to sense a ``chemical" gradient and climb it towards its maximum. To do so, we added a separate channel $\Gamma : \supp \to \R_{\ge 0}$ whose activations are defined following a Gaussian function around a point randomly sampled on a circle surrounding the center of the CA for each evaluation of the optimization process ensuring creatures learn to follow gradient and not a fixed direction while keeping the distance to cover constant. We also added 5 kernels and growth functions from $\Gamma$ to $A$, which are also optimized, so the creature is able to sense the chemical. The fitness of an individual is then computed with the following function :

\begin{equation*}
    f(\theta) = \frac{\sum_{x\in \mathcal{L}} A^{500}_{\Sigma}(x) \times \Gamma(x)}{\sum_{x \in \supp} A^{500}_{\Sigma}(x)}
\end{equation*}

Since mass is conserved, the optimal behavior for a creature is to concentrate as much of its mass in the cells where $\Gamma$ is maximal.\\
We have been able to find good solutions to this task as shown in figure \ref{fig:evo_results}. Best solutions such as the one shown in figure \ref{fig:evo_creas} (d) are perfectly able to climb the gradient towards its maximum.

\section{Flow Lenia : Towards intrinsic evolution}

\begin{figure}[ht!]
\begin{center}
    \begin{tabular}{cc}
    \multicolumn{2}{c}{\subfloat[]{\includegraphics[width=.95\columnwidth]{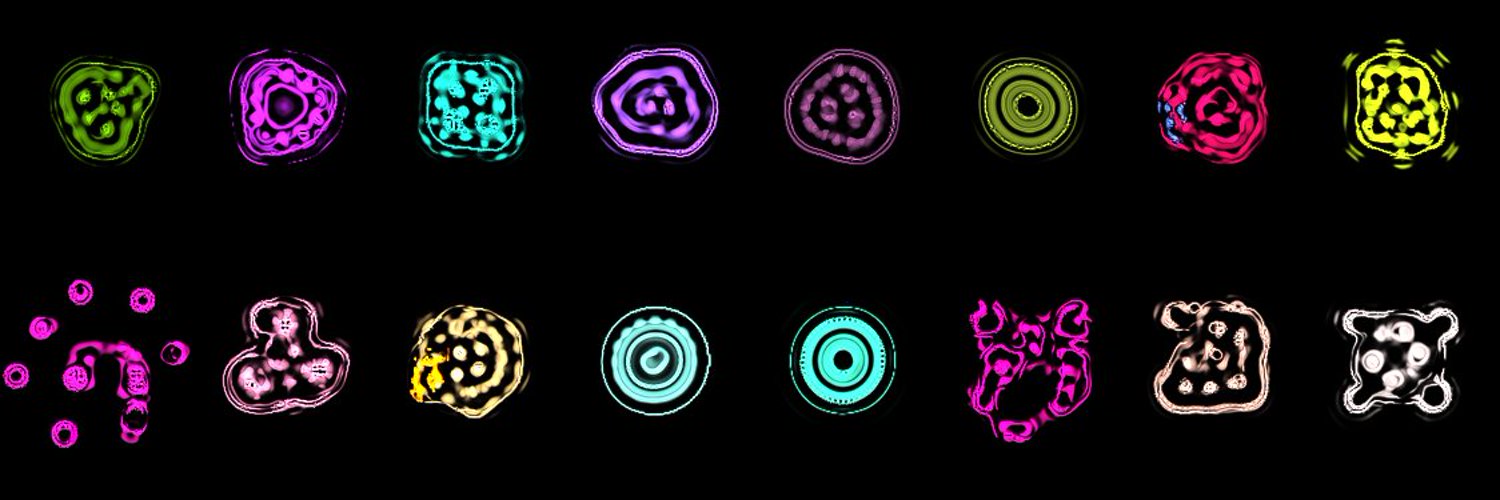}}}\\
    \subfloat[]{\includegraphics[width=0.45\linewidth]{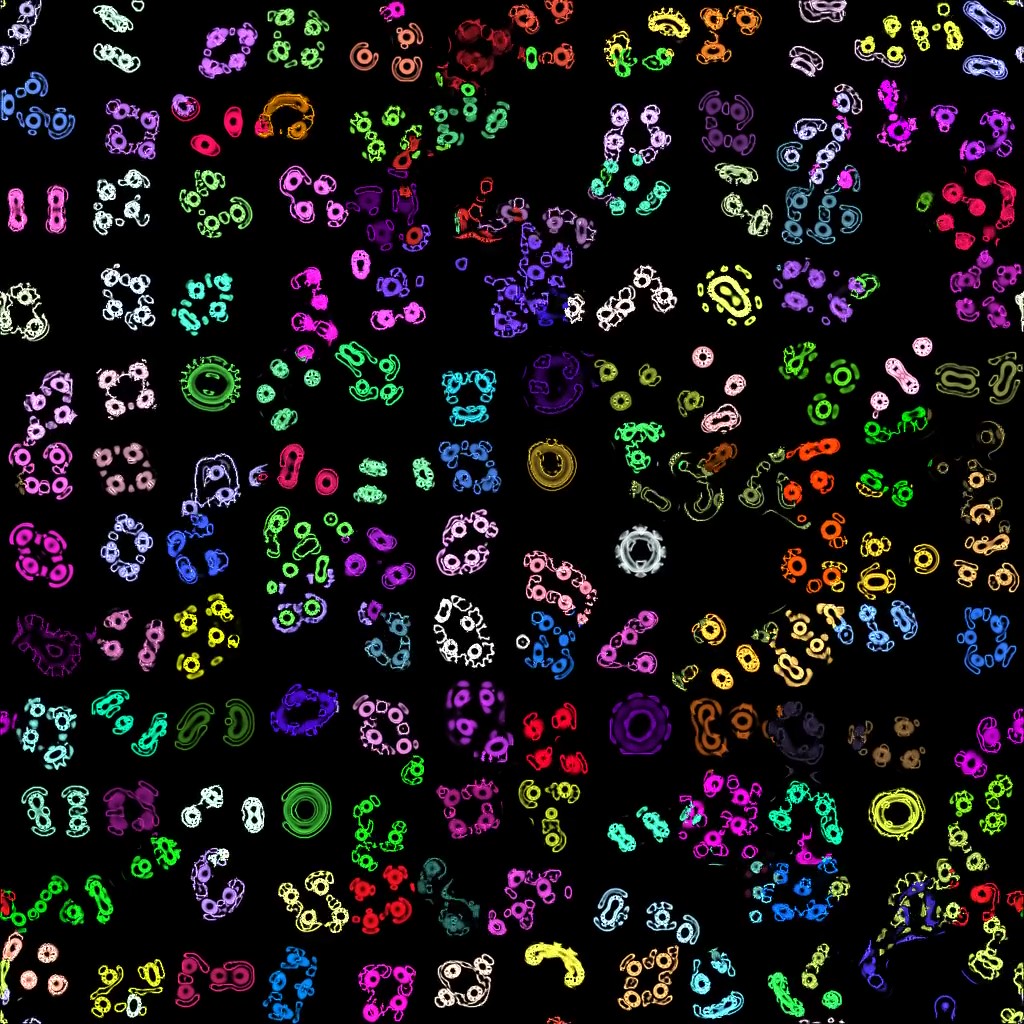}} &
    \subfloat[]{\includegraphics[width=0.45\linewidth]{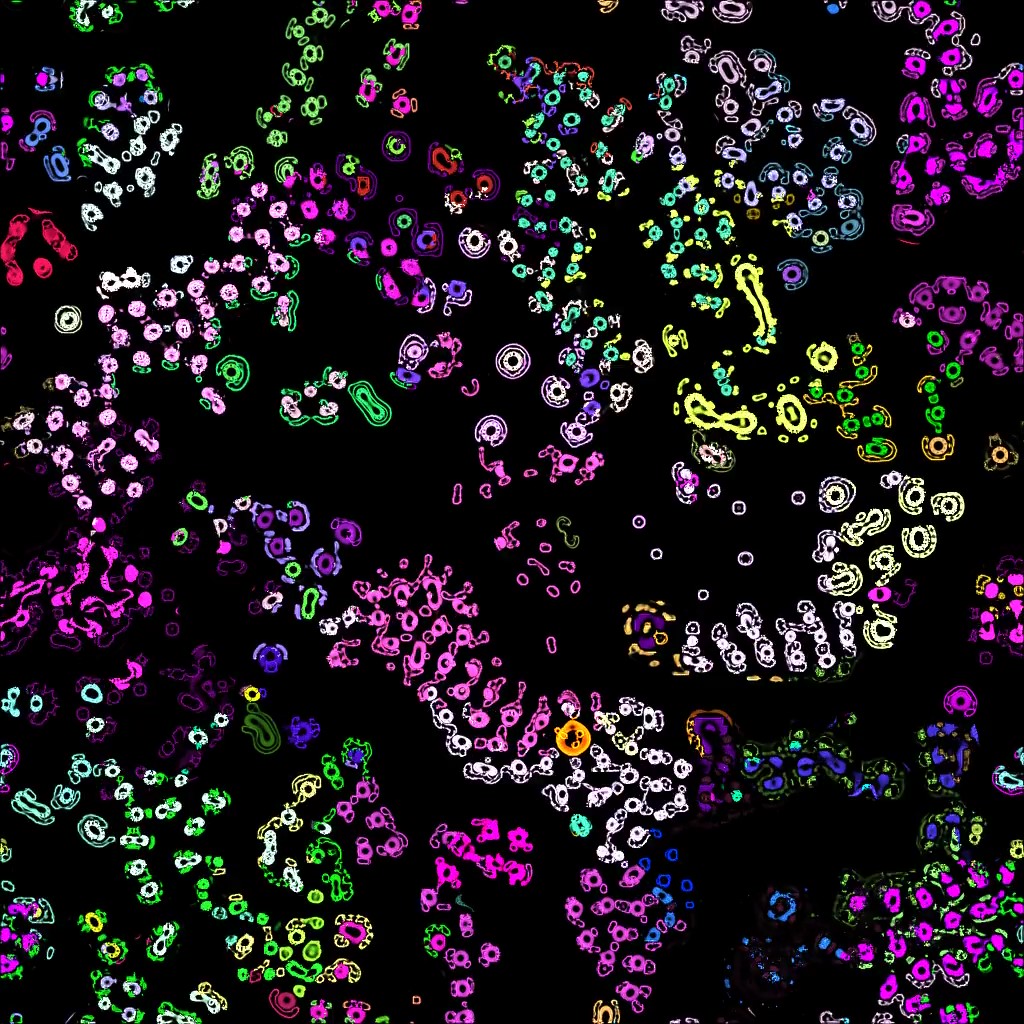}} \\
    \subfloat[]{\includegraphics[width=0.45\linewidth]{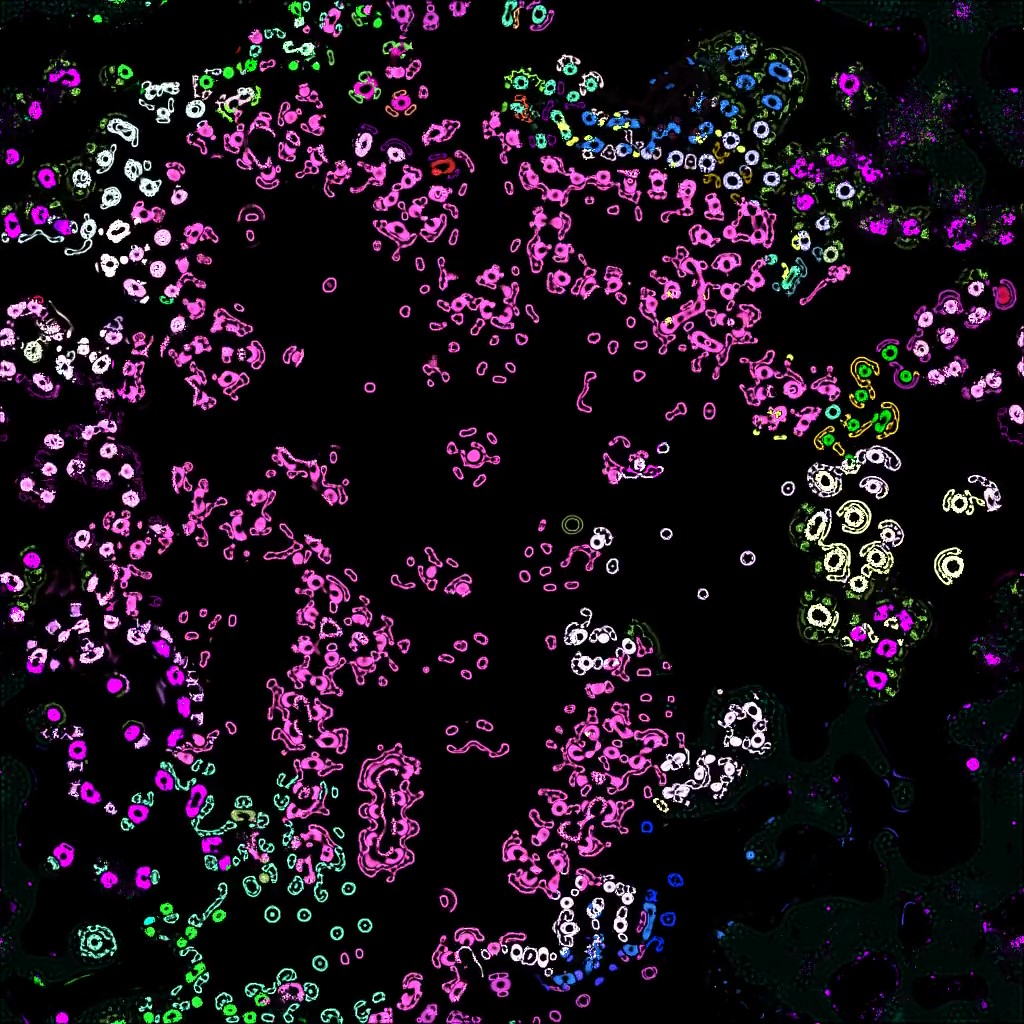}} &
    \subfloat[]{\includegraphics[width=0.45\linewidth]{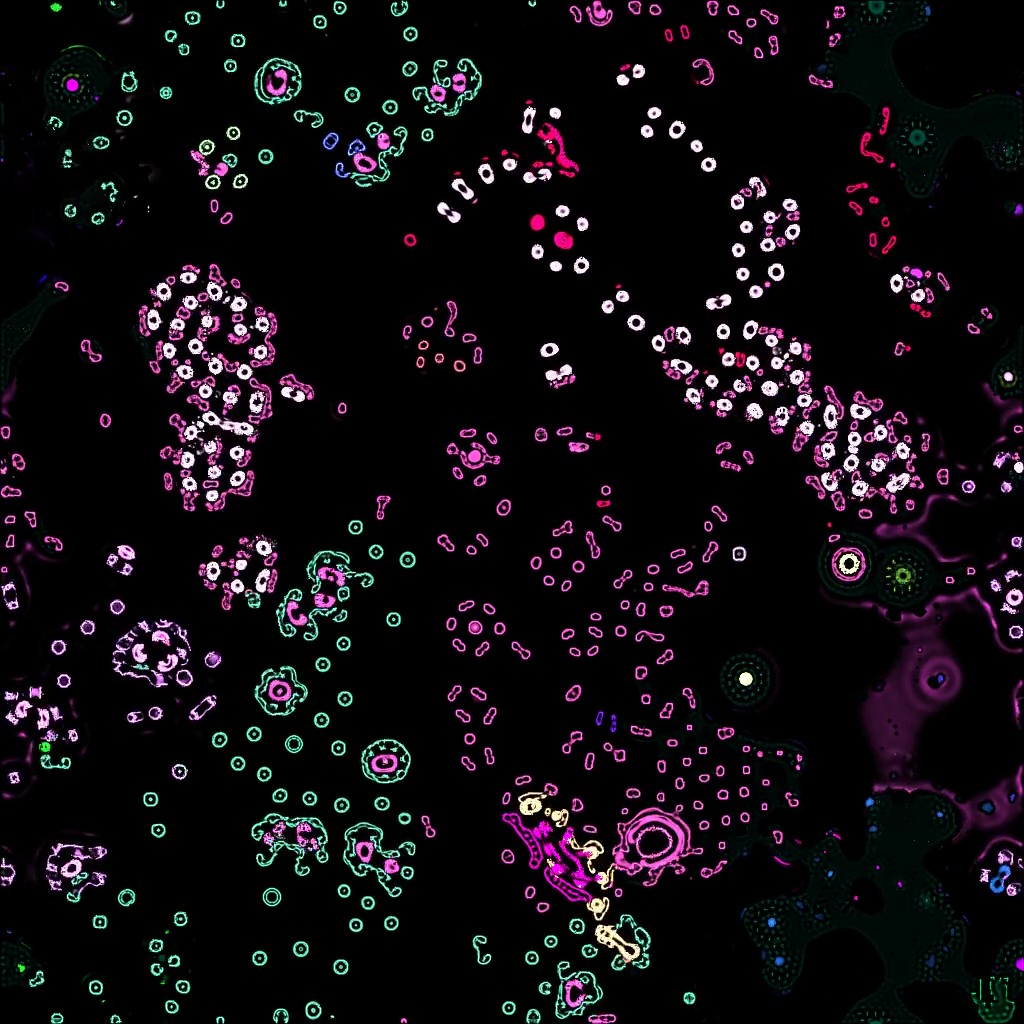}} \\
    \multicolumn{2}{c}{\subfloat[]{\includegraphics[width=.95\columnwidth]{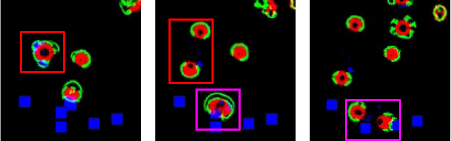}}}\\
    \end{tabular}
\end{center}
\caption{Multi-species simulations. (a) Sample of a multi-species simulation where color code for parameters. (b-e) Timelapse of larger scale multi-species simulation. World is a $1024 \times 1024$ grid initialized with $144$ creatures with distinct parameters represented by colors. We simulate 200k timesteps and use softmax sampling as the mixing rule and random mutations every $500$ steps. (f) Timelapse of simulation with parameter embedding and food (in blue) showing division events (highlighted with boxes). Videos are available at \website.}
\label{fig:large}
\end{figure}

As said in the introduction, Flow Lenia allows to embed the update rule parameters inside the CA. Intuitively, we can ``attach'' a vector of parameters to the matter locally modifying how it behaves (i.e locally modifying how the affinity map is computed), and let it flow with it. Formally, this comes to defining a parameter map $P : \mathcal{L} \to \Theta$ where $\Theta$ is a given parameter set. This map can be used to locally modify the update rule. For instance, we can embed the $h \in \mathbb{R}^{|K|}$ vector weighting the importance of each kernel in the affinity map computation (see equation \ref{PotEq}), giving :

\begin{equation}
    U^t_j(p) = \sum_{i, k}P^t_k(p) \cdot G_k(K_k \ast A^t_i)(p)
\end{equation}

Then parameters can be moved along with matter. A question is how to mix parameters arriving in the same cell. Here we propose two different methods which are respectively \emph{average} and \emph{softmax sampling}. The former makes a weighted average of incoming parameters with respect to the quantities of incoming matter and is formally defined as :

\begin{equation}
    P^{t+dt}(p) = \frac{\sum_{p' \in \mathcal{L}}A^t(p') I(p', p) P^t(p')}{\sum_{p' \in \mathcal{L}}A^t(p') I(p', p)}
\end{equation}

Softmax sampling on the other hand samples a parameter in the set of incoming ones following the softmax distribution given by incoming quantities of matter :

\begin{equation}
    \mathbb{P}[P^{t+dt}(p) = P^{t}(x)] = \frac{\exp(A^t(x) I(x, p))}{\sum_{p' \in \mathcal{L}}\exp( A^t(p') I(p', p))} 
\end{equation}

Intuitively, the more represented set of parameters has a greater probability of being selected in the cell, like simulating in one step a competition between different parameters in the cell. Note that we could also sample each element of the vector of parameters independently giving some crossover mechanism. We can also add mutations in the simulation. For instance, we can, at a given rate, modify parameters in a randomly sampled zone by adding Gaussian noise to the parameter map.\\ 
The parameters embedding mechanism allows to run multi-species simulations where all creatures patches are initialized with different parameters as shown in figure \ref{fig:large}(a) where colors code for parameters. We can see that the creatures, although in the same CA, display different morphologies even though they have been initialised with the same pattern. We also ran larger scale simulations (e.g $1024 \times 1024$ worlds initialized with 144 creatures and simulated for 200k timesteps) using parameter embedding as shown in figure \ref{fig:large}(b-e). Such a simulation display very interesting dynamics on the large scale where some ``species" can take over large parts of the world by contaminating other species before reaching more stable states. Also, we can see the emergence of coherent creatures composed of different parameters where one of those will compose a sort of membrane around a nuclei composed by different parameters (see green and pink creature on bottom left of figure \ref{fig:large}(e)). Such a pattern can be seen as a form of symbiosis. \\
While parameter embedding could lead to the emergence of intrinsic evolutionary processes where parameters would compete for available matter, we can also add intrinsic selective pressures in the form of resource collection mechanisms bootstrapping evolution. The idea is to add food resources that creatures would need to collect in order to replenish their own constantly decaying pool of resources. To do so, we let matter decay at a fixed rate $\rho_{decay}$, and create a food map $\Psi : \mathcal{L} \to \mathbb{R}_{\ge 0}$. When matter is in a cell where there is also food, then food is transformed into matter at a given rate $\rho_{digest}$ giving the following update.
\begin{equation}
    \begin{cases}
        A^{t+dt}(x) = \cdots + [A^t(x)\rho_{digest}]_0^{\Psi^t(x)} - A^t(x) \rho_{decay}\\
        \Psi^{t+dt}(x) = \Psi^t(x) - [A^t(x)\rho_{digest}]_0^{\Psi^t(x)}
    \end{cases}
\end{equation}
Where $\cdots$ refers to the update equation \ref{FlowApplyEq} and $[\cdot]_a^b$ is the clip function between $a$ and $b$.
We enable creatures to sense food by adding kernels and growth function from the food map $\Psi$ to creatures' channels $A$.\\
By performing simulations with random set of parameters, parameter embedding and food, we have been able to observe interesting patterns. First, we have been able to observe that some creatures, while not having trained for it, are able to go towards nearby food sources and consume it. We can hypothesize that creatures with such a capability will survive (and grow) while other will not leading to intrinsic evolution. Quite interestingly, complex patterns can emerge from the change of mass induced by decay or food consumption. For instance, when growing after eating, some creatures will divide in two identical creatures as shown in figure \ref{fig:large} (f), crucial pattern for evolution to occur. On the other hand, mass decay also leads to interesting dynamics where creatures undergo phase transitions, changing their shape and behavior, when their mass falls below a certain threshold which can lead them to adopt foraging behaviour for example while being initially static.

\section{Discussion}

Due to the mass conservative nature of Flow Lenia, most of patterns do not grow indefinitely into spatially global patterns (i.e patterns that diffuse on the entire grid also called Turing-like patterns), therefore SLPs are much more common and easier to find. This is an important difference from the previous versions of Lenia, where one needs to search or evolve for patterns that are both non-vanishing and non-exploding, and to constantly monitor their existential status. Here, the mass conservation constraint acts as a regularizer on the kind of patterns that can emerge.
Even though patterns generated by Flow Lenia are often static or slowly moving, we have been able to find creatures with complex dynamics from random search only which would be a difficult task in Lenia as most of the search space corresponds to either exploding or vanishing patterns. Furthermore, we have shown that the update rule parameters can be optimized with simple evolutionary strategies to generate patterns with specific properties and behaviors. Doing so in Lenia is a difficult task since the spatial localization of emergent patterns is not guaranteed necessitating more complex algorithms accounting for such a property.\\
Finally we showed that the Flow Lenia system allowed for the integration of the update rule parameters within the CA dynamics allowing for the coexistence of multiple update rules, and thus different creature or species, within the same simulation. We argue that such a feature represent an important step towards the design of emergent microcosms \citep{arbesman2022} in which could emerge intrinsic, maybe open-ended, evolutionary processes through inter-species interactions. Moreover, we also showed that food consumption mechanisms can be implemented in Flow Lenia which could act as intrinsic selective pressures in such settings bootstrapping evolutionary processes though competitive dynamics. Whereas environment design is poorly addressed and quite challenging in cellular automata systems, we believe that it is crucial to study the emergence of agency and cognition in those systems as argued in \citet{godfrey-smith2002} and shown in \citet{hamon:hal-03519319}. By enabling the design of complex environmental features like inter-species interactions, walls, food or temperature, Flow Lenia could represent a particularly interesting system to study ecological theories of the evolution of complexity and cognition \citep{nisioti2021}.\\
Lot of exciting roads remain to be taken in order to fully capture the value of complex self-organized systems such as Flow Lenia. A major path to explore is about the detection of agency and cognition in such systems where everything is emergent with no predefined notion of individuals. Information theoretical measures of concepts like individuality, autonomy and agency \citep{krakauer2020,bertschinger2008,negru2021,kolchinsky2018} could be of great interest in this perspective even though they remain difficultly applicable in large systems. 
Demonstrating the emergence of intrinsic evolutionary processes has been left out of the scope of this paper and left for future work. However, evaluating such a system appears as a difficult task. Quantifying intrinsic evolutionary processes with measures like evolutionary activity \citep{bedau1996,droop2012} could reveal to be useful in order to get a better grasp on the complex dynamics taking place in simulations such as the one shown in figure \ref{fig:large}. Ultimately, such simulations could be key for the growth of complexity inside the system and the emergence of virtual creatures displaying more complex forms of cognition.

\footnotesize
\bibliographystyle{apalike}
\bibliography{Lenia_flowers} 

\begin{thebibliography}{}

\bibitem[Arbesman, 2022]{arbesman2022}
Arbesman, S. (2022).
\newblock Emergent {{Microcosms}}.

\bibitem[Bedau and Packard, 1996]{bedau1996}
Bedau, M.~A. and Packard, N.~H. (1996).
\newblock Measurement of {{Evolutionary Activity}}, {{Teleology}}, and
  {{Life}}.

\bibitem[Bertschinger et~al., 2008]{bertschinger2008}
Bertschinger, N., Olbrich, E., Ay, N., and Jost, J. (2008).
\newblock Autonomy: {{An}} information theoretic perspective.
\newblock {\em Bio Systems}, 91(2):331--345.

\bibitem[Bradbury et~al., 2018]{jax2018github}
Bradbury, J., Frostig, R., Hawkins, P., Johnson, M.~J., Leary, C., Maclaurin,
  D., Necula, G., Paszke, A., VanderPlas, J., {Wanderman-Milne}, S., and Zhang,
  Q. (2018).
\newblock {{JAX}}: Composable transformations of {{Python}}+{{NumPy}} programs.

\bibitem[Chan, 2019]{chan2019}
Chan, B. W.-C. (2019).
\newblock Lenia - {{Biology}} of {{Artificial Life}}.
\newblock {\em Complex Systems}, 28(3):251--286.

\bibitem[Chan, 2020]{chan2020}
Chan, B. W.-C. (2020).
\newblock Lenia and {{Expanded Universe}}.
\newblock In {\em The 2020 {{Conference}} on {{Artificial Life}}}, pages
  221--229.

\bibitem[Droop and Hickinbotham, 2012]{droop2012}
Droop, A. and Hickinbotham, S. (2012).
\newblock A quantitative measure of non-neutral evolutionary activity for
  systems that exhibit intrinsic fitness.
\newblock In {\em {{ALIFE}} 2012: {{The Thirteenth International Conference}}
  on the {{Synthesis}} and {{Simulation}} of {{Living Systems}}}, pages 45--52.
  {MIT Press}.

\bibitem[Froese and Ziemke, 2009]{froese2009}
Froese, T. and Ziemke, T. (2009).
\newblock Enactive artificial intelligence: {{Investigating}} the systemic
  organization of life and mind.
\newblock {\em Artificial Intelligence}, 173(3):466--500.

\bibitem[{Godfrey-Smith}, 2002]{godfrey-smith2002}
{Godfrey-Smith}, P. (2002).
\newblock Environmental complexity and the evolution of cognition.
\newblock In {\em The Evolution of Intelligence}, pages 223--249. {Lawrence
  Erlbaum Associates Publishers}, {Mahwah, NJ, US}.

\bibitem[Hamon et~al., 2022]{hamon:hal-03519319}
Hamon, G., Etcheverry, M., Chan, B. W.-C., {Moulin-Frier}, C., and Oudeyer,
  P.-Y. (2022).
\newblock Learning sensorimotor agency in cellular automata.

\bibitem[Hickinbotham and Stepney, 2015]{hickinbotham2015}
Hickinbotham, S. and Stepney, S. (2015).
\newblock Conservation of {{Matter Increases Evolutionary Activity}}.
\newblock In {\em {{ECAL}} 2015: {{The}} 13th {{European Conference}} on
  {{Artificial Life}}}, pages 98--105. {MIT Press}.

\bibitem[Kingma and Ba, 2017]{kingma2017}
Kingma, D.~P. and Ba, J. (2017).
\newblock Adam: {{A Method}} for {{Stochastic Optimization}}.

\bibitem[Kolchinsky and Wolpert, 2018]{kolchinsky2018}
Kolchinsky, A. and Wolpert, D.~H. (2018).
\newblock Semantic information, autonomous agency and non-equilibrium
  statistical physics.
\newblock {\em Interface Focus}, 8(6):20180041.

\bibitem[Krakauer et~al., 2020]{krakauer2020}
Krakauer, D., Bertschinger, N., Olbrich, E., Flack, J.~C., and Ay, N. (2020).
\newblock The information theory of individuality.
\newblock {\em Theory in Biosciences}, 139(2):209--223.

\bibitem[Lange, 2022]{lange2022}
Lange, R.~T. (2022).
\newblock Evosax: {{JAX-based Evolution Strategies}}.

\bibitem[Lehman and Stanley, 2011]{lehman2011a}
Lehman, J. and Stanley, K.~O. (2011).
\newblock Novelty {{Search}} and the {{Problem}} with {{Objectives}}.
\newblock In Riolo, R., Vladislavleva, E., and Moore, J.~H., editors, {\em
  Genetic {{Programming Theory}} and {{Practice IX}}}, Genetic and
  {{Evolutionary Computation}}, pages 37--56. {Springer}, {New York, NY}.

\bibitem[Mordvintsev et~al., 2022]{mordvintsev2022}
Mordvintsev, A., Niklasson, E., and Randazzo, E. (2022).
\newblock Particle {{Lenia}} and the energy-based formulation.
\newblock
  https://google-research.github.io/self-organising-systems/particle-lenia/.

\bibitem[Moroz, 2020]{moroz2020}
Moroz, M. (2020).
\newblock Reintegration tracking.
\newblock https://michaelmoroz.github.io/Reintegration-Tracking/.

\bibitem[Negru, 2021]{negru2021}
Negru, T. (2021).
\newblock Self, {{Agency}} and {{Autonomy}} in {{Dynamical Living Systems}}.
\newblock {\em Synthesis philosophica}, 36(1):191.

\bibitem[Nisioti et~al., 2021]{nisioti2021}
Nisioti, E., {Jodogne-del Litto}, K., and {Moulin-Frier}, C. (2021).
\newblock Grounding an {{Ecological Theory}} of {{Artificial Intelligence}} in
  {{Human Evolution}}.
\newblock In {\em {{NeurIPS}} 2021 - {{Conference}} on {{Neural Information
  Processing Systems}} / {{Workshop}}: {{Ecological Theory}} of {{Reinforcement
  Learning}}}, {virtual event, France}.

\bibitem[Oros and Nehaniv, 2007]{oros2007}
Oros, N. and Nehaniv, C. (2007).
\newblock {\em Sexyloop: {{Self-Reproduction}}, {{Evolution}} and {{Sex}} in
  {{Cellular Automata}}}.

\bibitem[Rafler, 2011]{rafler2011}
Rafler, S. (2011).
\newblock Generalization of {{Conway}}'s "{{Game}} of {{Life}}" to a continuous
  domain - {{SmoothLife}}.

\bibitem[Salimans et~al., 2017]{salimans2017}
Salimans, T., Ho, J., Chen, X., Sidor, S., and Sutskever, I. (2017).
\newblock Evolution {{Strategies}} as a {{Scalable Alternative}} to
  {{Reinforcement Learning}}.
\newblock {\em arXiv:1703.03864 [cs, stat]}.

\bibitem[Sayama, 1999]{sayama1999}
Sayama, H. (1999).
\newblock Toward the realization of an evolving ecosystem on cellular automata.
\newblock {\em Proc. Fourth Int. Symp. Artificial Life and Robotics}, pages
  254--257.

\bibitem[Stanley, 2019]{stanley2019}
Stanley, K.~O. (2019).
\newblock Why {{Open-Endedness Matters}}.
\newblock {\em Artificial Life}, 25(3):232--235.

\end{thebibliography}

\end{document}